%% file: neurips_2026.tex
\documentclass{article}

\PassOptionsToPackage{numbers, compress}{natbib}

\usepackage[preprint]{neurips_2026}

\usepackage[utf8]{inputenc} 
\usepackage[T1]{fontenc}    
\usepackage{hyperref}       
\usepackage{url}            
\usepackage{booktabs}       
\usepackage{amsfonts}       
\usepackage{nicefrac}       
\usepackage{microtype}      
\usepackage{pifont}
\usepackage{graphicx}
\usepackage{multirow}
\usepackage{amsmath}
\usepackage[table]{xcolor}
\usepackage{float}
\usepackage{enumitem}

\hypersetup{
	colorlinks=true,
	urlcolor=blue,
}

\title{WorldVLN: Autoregressive World Action Model for Aerial Vision-Language Navigation}

\author{%
  Baining Zhao$^{1}$\thanks{All authors contributed equally to this research.},
  Jiacheng Xu$^{2}$\footnotemark[1],
  Weicheng Feng$^{2}$\footnotemark[1],
  Xin Zhang$^{3}$\footnotemark[1],
  Zhaolu Wang$^{4}$\footnotemark[1], \\
  \textbf{Haoyang Wang$^{1}$, 
  Shilong Ji$^{1}$, 
  Ziyou Wang$^{5}$, 
  Jianjie Fang$^{1}$, 
  Zhiheng Zheng$^{1}$,} \\
  \textbf{Weichen Zhang$^{1}$, 
  Yu Shang$^{1}$, 
  Wei Wu$^{3}$, 
  Chen Gao$^{1}$\thanks{Corresponding authors.}, 
  Xinlei Chen$^{1}$\footnotemark[2], 
  Yong Li$^{1}$}\\
  $^{1}$Tsinghua University,
  $^{2}$Shandong University,
  $^{3}$Manifold AI, \\
  $^{4}$Beijing Institute of Technology,
  $^{5}$Northeastern University \\
  \texttt{zbn22@mails.tsinghua.edu.cn, chgao96@gmail.com, } \\
  \texttt{chen.xinlei@sz.tsinghua.edu.cn, liyong07@tsinghua.edu.cn}
}

\begin{document}

\maketitle

\begin{abstract}
Aerial vision-language navigation (VLN) requires agents to follow natural-language instructions through closed-loop perception and action in 3D environments. We argue that aerial VLN can be formulated as a prediction-driven world-action problem: the agent should anticipate latent world evolution and act according to the predicted consequences. To this end, we propose WorldVLN, the first autoregressive world action model for aerial VLN. Unlike full-sequence video-generation world models that generate an entire visual clip, WorldVLN adapts a latent autoregressive video backbone to predict short-horizon world-state transitions and directly decodes them into executable waypoint actions. After each action segment is executed, newly received observations are encoded back into the autoregressive context, enabling closed-loop world-action prediction. We further introduce a two-stage training framework that first grounds the video prior in instruction-conditioned navigation dynamics and then develops Action-aware GRPO, the first reinforcement learning method tailored to autoregressive WAMs, to optimize waypoint decisions through their downstream rollout consequences. On public outdoor and indoor benchmarks, WorldVLN consistently outperforms existing Vision-Language-Action baselines with 12\%+ success-rate gains and larger advantages on challenging cases. It further transfers zero-shot to real drone deployment, suggesting that the proposed WorldVLN offers a promising route for spatial action tasks. Demos and code are available at \url{https://embodiedcity.github.io/WorldVLN/}.

\end{abstract}

\section{Introduction}

Vision-Language Navigation (VLN) is one of the core tasks of spatial intelligence, where an agent follows human instructions and autonomously moves through a 3D environment~\cite{anderson2018vision,krantz2020beyond,wu2024vision}. The agent must understand high-level language, perceive partial egocentric observations, and progressively generate low-level actions in a closed-loop manner as new observations become available~\cite{krantz2020beyond,shah2023lm,song2025towards}, making generalizable VLN agents difficult to build. The rapid progress of foundation models, especially LLMs and VLMs, has created new opportunities to transfer general-purpose capabilities to embodied navigation~\cite{zhou2024navgpt,zhang2025citynavagent,qi2025vlnr1}. Following the Bitter Lesson~\cite{sutton2019bitter}, Vision--Language--Action (VLA) models extend vision-language models with action outputs and directly map observations and instructions to control commands~\cite{brohan2023rt1,kim2024openvla,zhang2024navid}. However, VLA models still suffer from limited generalization in embodied navigation, because their web-scale visual-linguistic priors are well suited for recognizing objects and parsing instructions, but not for modeling how the world evolves under the agent's own actions. As a result, they remain limited in capturing the temporal, geometric, and causal structure required for embodied action generation, treating embodied behavior as a conditional \textbf{mapping} from instruction and observation to action.

Actually, biological spatial intelligence~\cite{epstein2017cognitive,crivelli2023goal,erdem2012goal} suggests that navigation is inherently anticipatory: humans implicitly predict the state consequences of their own movements and select actions that are expected to bring the resulting state closer to the intended goal. Recent advances in visual foundation models, especially video generation models~\cite{yang2025cogvideoxtexttovideodiffusionmodels,wan2025wanopenadvancedlargescale}, have revealed the emergence of powerful predictive capabilities from large-scale visual-temporal pretraining~\cite{kang2025farvideogenerationworld,kim2026cosmospolicyfinetuningvideo}. Extending this direction, video-based world models learn how visual scenes evolve under action-conditioned dynamics and thereby acquire rich spatiotemporal priors over motion, viewpoint transitions, and physical evolution---precisely the structure that VLM-based VLA models lack. This observation reveals a different formulation of VLN as a \textbf{prediction} problem: given observations and instructions, the agent predicts how the world will evolve under candidate movements and selects the action whose anticipated consequences best satisfy the instructed goal.

However, realizing spatial action with existing video generation models remains nontrivial. A direct approach~\cite{chen2025largevideoplannerenables,emami2025diffusionmodelssmarteruavs} is to condition a video generation model on the VLN instruction and current observation, synthesize future visual observations, and then recover actions through visual odometry. Yet this pipeline exposes fundamental mismatches between video generation and embodied navigation, both in model structure and in the learning objective that shapes the underlying representations:
\begin{itemize}[leftmargin=*]
    \item \textbf{Model Structure:} Most video generation backbones~\cite{wu2025hunyuanvideo15technicalreport,chen2023videocrafter1opendiffusionmodels} generate an entire clip in a bidirectional manner, whereas embodied navigation requires a causal observe--act--update loop grounded in past and current observations. This mismatch is especially critical in aerial VLN, where large viewpoint changes and accumulated state errors require persistent memory and closed-loop correction.
    \item \textbf{Learning Objective:} Generic video generation models are optimized for visually plausible synthesis~\cite{NEURIPS2022_39235c56,ho2022imagenvideohighdefinition}, whereas VLN requires action-aware consequence modeling: the learned representations must not only predict how observations evolve, but also encode which state transitions are geometrically consistent, action-decodable, and beneficial for reaching the instructed goal.
\end{itemize}

To address these challenges, we propose \textbf{WorldVLN}, an autoregressive world action model (WAM) for aerial VLN, together with a two-stage training framework that aligns a video-generation backbone with world-action dynamics. 
\textbf{First}, WorldVLN repurposes a pre-trained video latent autoregressive transformer for closed-loop navigation. The backbone predicts short-horizon latent world transitions from the instruction and observation history, decodes them into waypoint actions via a designed action decoder, and feeds newly observed states back into the autoregressive context after execution.
\textbf{Then}, we train WorldVLN with a two-stage framework. Stage 1 uses supervised training to ground the video prior in instruction-conditioned navigation dynamics and train the action decoder to recover expert waypoint actions from latent world transitions. Stage 2 introduces \textbf{Action-aware Group Relative Policy Optimization
(GRPO)}, which performs online autoregressive rollouts and optimizes segment-level action decisions with trajectory, task, and reference rewards. A temporal decay weighting further emphasizes early decisions, encouraging the model to account for how current actions influence downstream observations, future actions, and final navigation success.
\textbf{Finally}, experiments on public indoor and outdoor UAV benchmarks show that WorldVLN significantly outperforms VLA baselines. We further explore three questions---whether WAM learns more effectively than VLA, why autoregressive prediction is necessary, and what Action-aware GRPO contributes---highlighting the effectiveness of the proposed architecture and training algorithm. WorldVLN also shows zero-shot transfer to real UAV deployment.
Our main contributions are summarized as follows:
\begin{itemize}[leftmargin=*]
    \item To our knowledge, we propose the \textbf{first autoregressive world action model for aerial VLN}, which temporally predicts latent world representations, directly decodes low-level navigation actions, and closes the loop by grounding subsequent decisions in newly received visual observations.

    \item We introduce the \textbf{first Action-aware GRPO method tailored to autoregressive WAMs}. After supervised navigation grounding, our Action-aware GRPO further aligns latent world-action representations with navigation outcomes.

    \item We achieve state-of-the-art results on both outdoor and indoor challenging aerial VLN benchmarks, and demonstrate zero-shot generalization on a real-world drone platform.
\end{itemize}

\section{Related work}

\paragraph{Vision-language-action models.}
VLA models extend pretrained vision-language models with action heads, enabling end-to-end mapping from language instructions and visual observations to executable actions~\cite{kim2024openvla,zhang2024navid,wang2026vla}. This paradigm has been widely explored in embodied control, including robotic manipulation~\cite{brohan2023rt1,intelligence2026pi07steerablegeneralistrobotic} and navigation~\cite{gao2026openflycomprehensiveplatformaerial,xue2026omninavunifiedframeworkprospective}, by transferring semantic priors from large-scale vision-language pretraining. However, their generalization in embodied navigation remains limited, because VLM-based VLAs primarily inherit priors for object recognition, instruction parsing, and scene understanding, but do not explicitly model action-conditioned world dynamics.

\paragraph{World action models.}
Video generation foundation models~\cite{yang2025cogvideoxtexttovideodiffusionmodels,wan2025wanopenadvancedlargescale,wu2025hunyuanvideo15technicalreport} provide strong visual-temporal priors over motion, viewpoint changes, and scene evolution, making them promising world modeling backbones~\cite{zhang2025epona,qin2024worldsimbenchvideogenerationmodels}. However, they are primarily optimized for realistic future synthesis rather than goal-directed action generation. Recent navigation methods often use world models in an imagine-and-rank manner, where multiple candidate routes are visually rolled out and then selected according to predicted outcomes~\cite{bar2025navigation,zhang2025aerial}. Although effective, this paradigm is indirect and computationally expensive. WAMs offer a more integrated alternative by coupling latent world prediction with action generation, either by recovering actions from predicted futures or directly decoding actions from world representations~\cite{kim2026cosmospolicyfinetuningvideo,pai2025mimicvideovideoactionmodelsgeneralizable}. Autoregressive WAMs~\cite{li2026causalworldmodelingrobot,ye2026worldactionmodelszeroshot} further support temporal memory and closed-loop feedback, but remain underexplored for spatial navigation, especially aerial VLN, where large viewpoint changes, continuous 3D motion, and accumulated state errors make autoregressive updating particularly important. Moreover, existing autoregressive WAMs often adapt bidirectional diffusion backbones via teacher- or self-forcing, incurring high computational cost and hindering direct optimization under the native observe--act--update interface.

\paragraph{Post-training methods.}
Post-training adapts pretrained foundation models to downstream task objectives. Current VLA and WAM policies are still largely trained by supervised fine-tuning on expert demonstrations, with WAMs sometimes using imagined observations or video prediction as additional supervision. Such imitation-based training fits the demonstration distribution but remains vulnerable to covariate shift and accumulated errors~\cite{ross2011reduction,mandlekar2021matters}. Reinforcement learning can directly optimize task rewards beyond demonstration likelihood~\cite{kumar2020conservative,chebotar2023q,shao2024deepseekmath}, but RL for autoregressive WAMs~\cite{chen2021decision} remains underexplored. Applying RL to autoregressive WAMs raises new challenges, including the mismatch between visually plausible world generation and action-outcome optimization, as well as credit assignment in multi-step closed-loop rollouts~\cite{hafner2023mastering,wu2023daydreamer}.

\section{Problem formulation}

We formulate aerial VLN as a partially observable sequential decision-making problem. Given a natural-language instruction $\ell$ at the starting position, the agent executes a sequence of actions to progressively complete the instruction in a 3D environment. Let $\pi_\theta$ denote the navigation policy. At step $t$, the agent receives an egocentric observation $o_t$ and predicts a waypoint action $a_t$ conditioned on the instruction, the current and historical observations, and the executed action history:
\begin{equation}
a_t \sim \pi_\theta(\cdot \mid o_{\le t}, a_{<t}, \ell),
\qquad
a_t = (\Delta x_t,\Delta y_t,\Delta z_t,\Delta \psi_t) \in \mathbb{R}^4 ,
\end{equation}
where $(\Delta x_t,\Delta y_t,\Delta z_t)$ represents the relative 3D translation and $\Delta \psi_t$ represents the relative yaw change.
Executing $a_t$ updates the agent pose $q_t=(x_t,y_t,z_t,\psi_t)$ and induces a new observation:
\begin{equation}
q_{t+1}=q_t\oplus a_t, \qquad o_{t+1}=\Omega(q_{t+1}),
\end{equation}
where $\oplus$ applies the relative waypoint displacement to the current pose, and $\Omega$ maps the updated pose to the corresponding egocentric visual observation in the 3D environment. The agent repeats this observe--act process until it predicts a stop action or reaches the maximum horizon. The navigation is considered successful if the final position $(x_T,y_T,z_T)$ is within a threshold $\epsilon$ of the ground-truth target position $(x^\star,y^\star,z^\star)$:
$\left\|(x_T,y_T,z_T)-(x^\star,y^\star,z^\star)\right\|_2 < \epsilon$.

\section{Method}

\begin{figure}[t]
  \centering
  \includegraphics[width=0.9\linewidth]{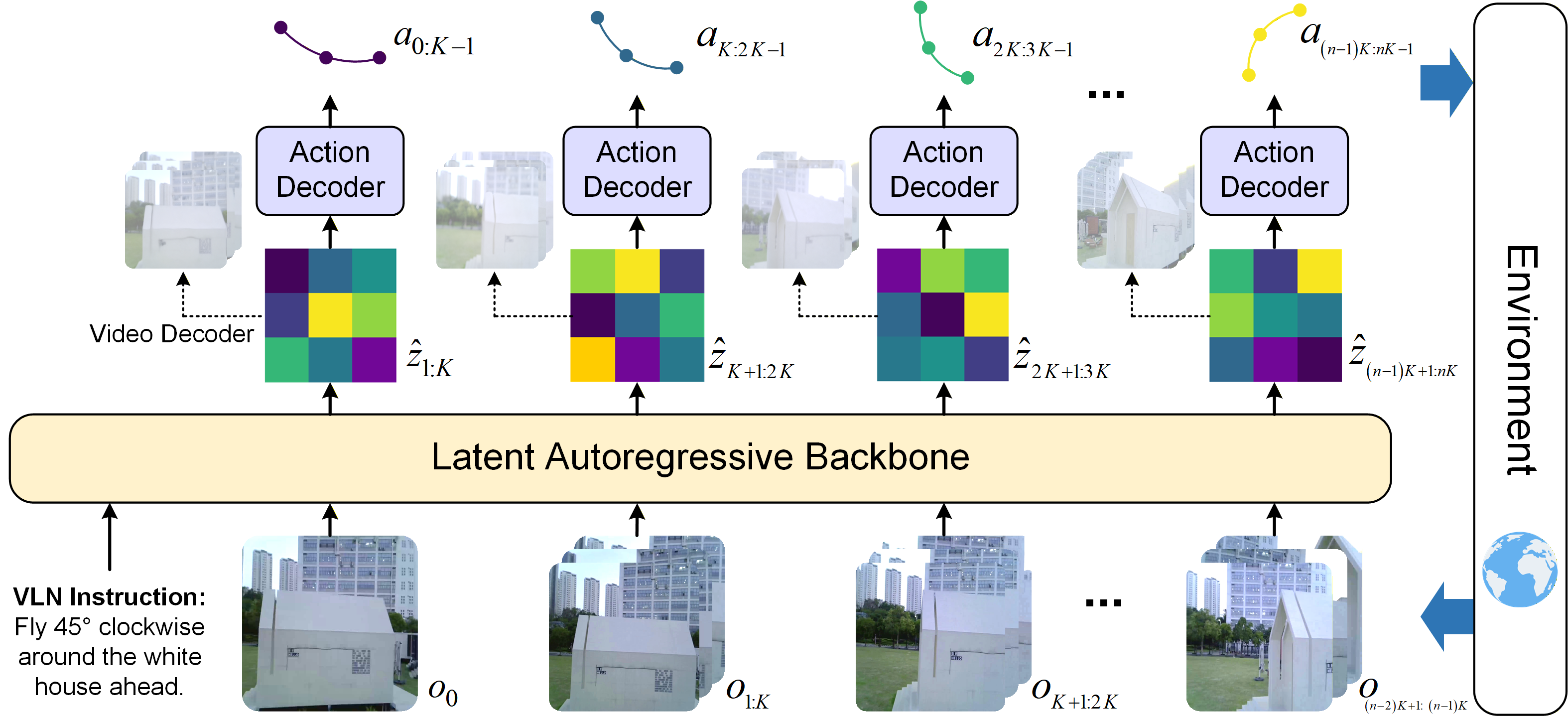}
  \caption{\textbf{WorldVLN architecture.} The model predicts short-horizon latent world transitions from the instruction and observation history, decodes them into waypoint actions, and updates the \textbf{autoregressive} context with newly observed states after execution. See Appendix~\ref{app:model_architecture} for details.}
  \label{fig:architecture}
\end{figure}

We first introduces the WorldVLN architecture for autoregressive world-action prediction, and then describes the two-stage training framework that combines supervised grounding with Action-aware GRPO.

\subsection{Model architecture}

As shown in Figure~\ref{fig:architecture}, WorldVLN adopts a pre-trained latent autoregressive video transformer as the world backbone to capture the temporal evolution of the agent's embodied state in latent space. Rather than treating the predicted latent as a video segment to be rendered, WorldVLN reinterprets it as a short-horizon world-state transition induced by the agent's movement, which provides the basis for action prediction.

The backbone consists of a text encoder $\psi$, a video VAE encoder $\mathcal{E}_{\mathrm{vid}}$, and a latent autoregressive Transformer $p_\theta$.
Let $e_\ell=\psi(\ell)$ be the encoded instruction, let $K$ denote the prediction horizon, and let $z_{\le t}$ denote the latent context encoded from real egocentric observations up to time $t$. Following the temporal autoregressive structure of the video backbone, the next latent world segment is predicted as
\begin{equation}
\hat{z}_{t+1:t+K}
\sim
p_\theta\left(\cdot \mid e_\ell, z_{\le t}\right).
\end{equation}
In the original video generation setting, $\hat{z}_{t+1:t+K}$ would be decoded into future frames and then used as part of the context for subsequent generation. In WorldVLN, we instead feed $\hat{z}_{t+1:t+K}$ to the action decoder $D_\phi$:
\begin{equation}
a_{t:t+K-1}=D_\phi\left(\hat{z}_{t+1:t+K}\right).
\end{equation}
After executing $a_{t:t+K-1}$, the agent receives real egocentric observations $o_{t+1:t+K}$, which are encoded into real latents:
\begin{equation}
z_{t+1:t+K}=\mathcal{E}_{\mathrm{vid}}\left(o_{t+1:t+K}\right).
\end{equation}
Instead of continuing generation with the model-predicted latent $\hat{z}_{t+1:t+K}$, WorldVLN replaces it with the real latent $z_{t+1:t+K}$ in the autoregressive context. The resulting closed-loop rollout is:
\begin{equation}
(e_\ell,z_{0})
\rightarrow \hat{z}_{1:K}
\rightarrow a_{0:K-1}
\rightarrow o_{1:K}
\rightarrow z_{1:K}
\rightarrow \hat{z}_{K+1:2K}
\rightarrow \cdots .
\end{equation}
Thus, each generated latent segment is used to decode a waypoint action sequence, while subsequent autoregressive prediction is grounded in the real latent encoded from the actual observation segment.

\subsection{Training framework}

\begin{figure}[t]
  \centering
  \includegraphics[width=\linewidth]{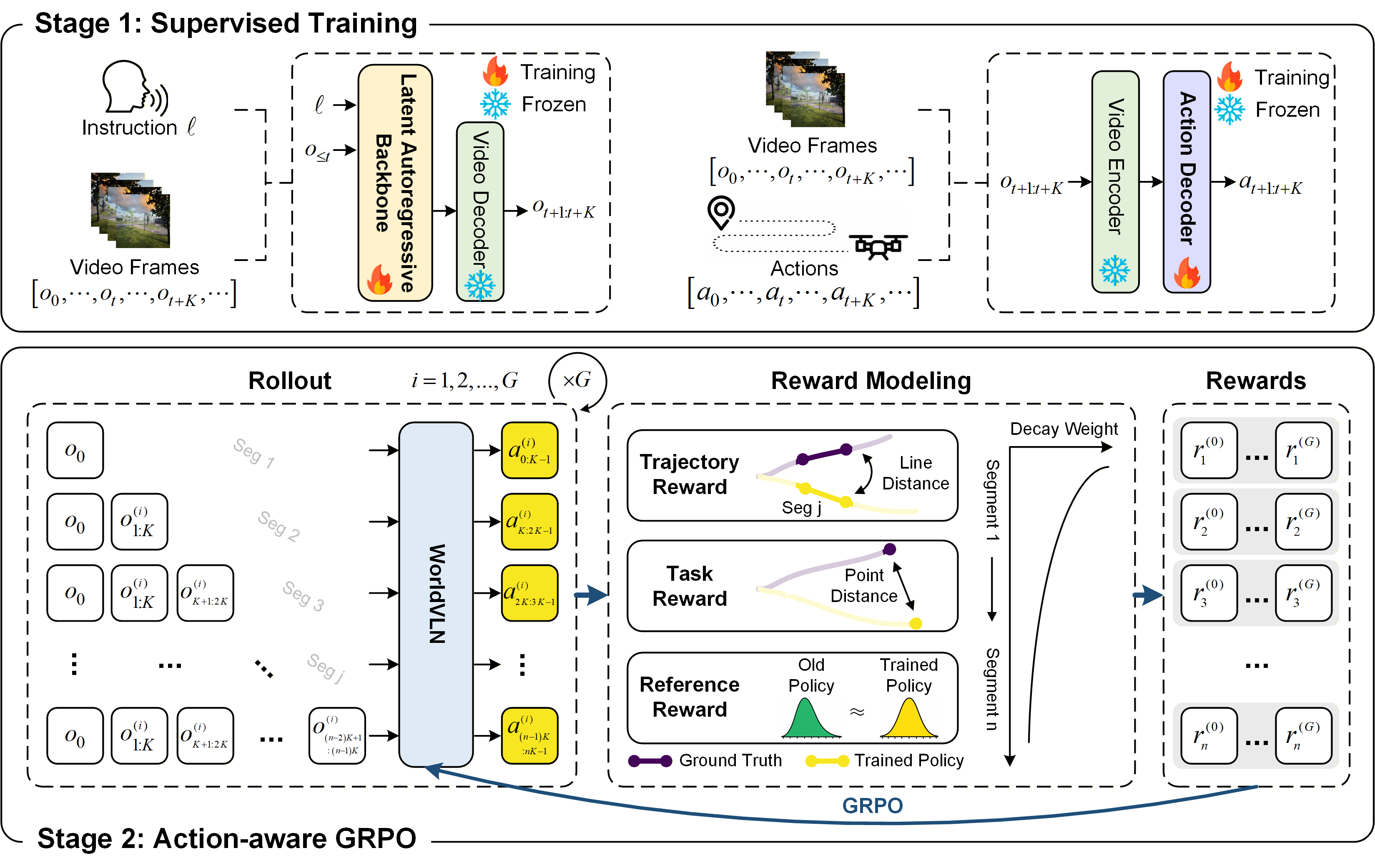}
  \caption{\textbf{Training framework}. Stage 1 supervises the latent autoregressive backbone with instruction-video pairs and the action decoder with video-trajectory pairs. Stage 2 samples multiple rollouts, assigns segment-level rewards from trajectory accuracy, task progress, and reference-policy regularization with temporal decay weighting, and updates WorldVLN through \textbf{Action-aware GRPO}.}

  \label{fig:training}
\end{figure}

We train the autoregressive WAM in two stages, as shown in Figure~\ref{fig:training}. Stage 1 uses supervised training to ground the video prior in instruction-conditioned navigation dynamics and make latent world representations action-decodable. Stage 2 introduces \textbf{Action-aware GRPO}, which uses online rollout rewards to optimize waypoint decisions according to their navigation consequences.

\subsubsection{Stage 1: Supervised training}

We train the backbone and the action decoder with supervised objectives, respectively. For the latent autoregressive backbone, we place the model back into its original autoregressive video-generation formulation, but use paired navigation instructions and egocentric navigation videos as training data. We encode the instruction $\ell$ as $e_\ell=\psi(\ell)$, and divide the corresponding video into $n+1$ observation segments $\{o_0, o_{1:K}, \ldots,o_{(n-1)K+1:nK}\}$. Each segment is encoded into a latent representation $z_{t+1:t+K}=\mathcal{E}_{\mathrm{vid}}\left(o_{t+1:t+K}\right), \quad
\text{for } t = 0, K, \ldots, (n-1)K
$. Following the temporal autoregressive objective of the backbone, we train it to predict each ground-truth future latent segment from the instruction and previous ground-truth latent context:
\begin{equation}
\mathcal{L}_{\mathrm{wm}}
=
-\sum
\log p_\theta\left(z_{t+1:t+K} \mid e_\ell, z_{\le t}\right).
\end{equation}

For the action decoder, we use paired navigation videos and trajectories as training data. Each video and trajectory are similarly divided into segments $\{o_{t+1:t+K},a^*_{t:t+K-1}\}$. We encode each video segment as $z_{t+1:t+K}$ and train the action decoder to recover the expert action. This matches the final inference interface, where the decoder receives a latent world transition and outputs executable waypoint actions. To accelerate convergence, we initialize the decoder with features from the video decoder and a learning-based visual odometry backbone, as both provide useful priors for mapping visual state transitions to camera-pose motion. The action decoder is optimized by
\begin{equation}
\mathcal{L}_{\mathrm{act}}
=
\sum
\left\|
D_\phi \left(\mathcal{E}_{\mathrm{vid}}\left(o_{t+1:t+K}\right) \right)-a^*_{t:t+K-1}
\right\| .
\end{equation}

\subsubsection{Stage 2: Action-aware GRPO}

To further align the autoregressive WAM with navigation outcomes, we introduce action-aware GRPO. During training, the model performs online autoregressive rollouts in the simulator, following the same observe--act process used at inference time. Given an instruction and the initial observation, the model predicts a latent segment, decodes waypoint actions, executes them in the environment, receives new observations, and repeats this process until the end of the navigation case.

For each navigation case, a group of $G$ online rollouts is sampled from the current policy and each rollout contains $n$ autoregressive decision segments. Suppose the $j$-th action segment in the $i$-th rollout is denoted by $a^{(i)}_{(j-1)K:jK-1}$. We assign reward to it by:

\begin{equation}
r^{(i)}_j
= \gamma^{j-1} \left(
\lambda_{\mathrm{traj}} r_{\mathrm{traj},j}^{(i)}
+
\lambda_{\mathrm{task}} r_{\mathrm{task},j}^{(i)}
+
\lambda_{\mathrm{ref}} r_{\mathrm{ref},j}^{(i)} \right).
\end{equation}

\textbf{Trajectory reward}
provides local geometric supervision by measuring how closely the predicted action $a$ follows the expert action $a^\star$. Since each segment consists of multiple waypoints, we compute the trajectory distance and convert it into a reward:
\begin{equation}\label{eq:trajectory_reward}
r_{{\rm{traj}},j}^{(i)} = \frac{1}{{1 + \left\| {a_{(j - 1)K:jK - 1}^{(i)} - a_{(j - 1)K:jK - 1}^{ \star (i)}} \right\|}}.
\end{equation}

\textbf{Task reward}
provides global outcome evaluation by evaluating how the autoregressive rollout induced by local actions affects final goal reaching. We compute the terminal distance between the rollout endpoint and the ground-truth target, and assign a higher reward to goal-reaching outcomes:
\begin{equation}
r_{{\rm{task}},j}^{(i)} = \frac{1}{{1 + {{\left\| {(x_T^{(i)},y_T^{(i)},z_T^{(i)}) - ({x^{ \star (i)}},{y^{ \star (i)}},{z^{ \star (i)}})} \right\|}_2}}}.
\end{equation}

\textbf{Reference reward} regularizes the updated policy toward the reference policy to prevent excessive policy drift. This term helps preserve the implicit imagination capability learned during supervised world-action training, preventing the latent world prediction from drifting too far away from the original navigation dynamics prior. We evaluate the probability of the sampled segment action under the reference policy $\pi_{\mathrm{ref}}$:
\begin{equation}
 r_{\mathrm{ref},j}^{(i)}
=
\log \pi_{\mathrm{ref}}\left(a_{(j - 1)K:jK - 1}^{(i)} \mid o_{\le t}, a_{<(j - 1)K}^{(i)}, \ell \right).
\end{equation}

\textbf{Decay weighting} $\gamma^{j-1},0<\gamma<1$ is integrated to reflect the asymmetric influence of errors in autoregressive navigation. It enables earlier decisions to receive larger weights, as they affect a longer chain of future observations and actions.

Finally, given a group of $G$ sampled rollouts, we compute the segment-level advantage by normalizing the rewards across the group, i.e., $A_j^{(i)}=(r_j^{(i)}-\mu_j)/(\sigma_j+\epsilon)$, where $\mu_j$ and $\sigma_j$ are the group statistics of $\{r_j^{(i)}\}_{i=1}^{G}$ for the $j$-th decision segment. We then update the policy induced by the autoregressive backbone and action decoder with the clipped GRPO objective:
\begin{equation}
\mathcal{J}_{\mathrm{GRPO}}
=
\mathbb{E}_{i,j}
\left[
\min
\left(
\rho_j^{(i)} A_j^{(i)},
\operatorname{clip}
\left(
\rho_j^{(i)},1-\epsilon_{\mathrm{clip}},1+\epsilon_{\mathrm{clip}}
\right)
A_j^{(i)}
\right)
\right],
\end{equation}
where $\rho_j^{(i)}=\pi_\theta(a_{(j-1)K:jK-1}^{(i)}\mid h_j^{(i)})/\pi_{{\mathrm{old}}}(a_{(j-1)K:jK-1}^{(i)}\mid h_j^{(i)})$, and $h_j^{(i)}$ denotes the rollout history before the $j$-th segment in the $i$-th rollout. By optimizing rewards computed from actual online rollouts, Action-aware GRPO provides direct supervision on action consequences and trains the model to account for how current waypoint decisions influence downstream observations, future actions, and final navigation success. See Appendix~\ref{subsec:training_framework} for details.

\section{Experiments}

We evaluate WorldVLN on both outdoor and indoor UAV benchmarks to verify its effectiveness across diverse aerial navigation scenarios. We compare it with representative VLN and VLA baselines, analyze training curves to highlight the potential of the WAM paradigm, and conduct ablations on the autoregressive architecture and Action-aware GRPO. Finally, we deploy WorldVLN on a real UAV platform to examine its generalization to real-world environments. 

\textbf{Experimental setup:} 
We evaluate WorldVLN on UAV-Flow~\cite{wang2025uav} and IndoorUAV~\cite{liu2026indooruav}, comparing it with the VLA baselines under the corresponding benchmark protocols. WorldVLN uses InfinityStar~\cite{liu2025infinitystar} as the latent autoregressive backbone, with the action decoder initialized from Wan VAE~\cite{wan2025wan} and TSformer-VO-style~\cite{franccani2025transformer} priors. Training is conducted on 8 NVIDIA A800 80GB GPUs, and simulator rollouts are executed on an RTX 4090 workstation. See Appendix~\ref{subsec:training_framework} and Appendix~\ref{subsec:exp_setup} for details.

\subsection{Quantitative results}

\begin{table*}[t]
	\centering
	\caption{Success Rates (SR, \%) on the UAV-Flow-Sim test set. See Appendix~\ref{subsec:uavflow_details}. }
	\label{tab:uavflow_results}
    \setlength{\tabcolsep}{2pt} 
	\resizebox{\linewidth}{!}{
		\begin{tabular}{l|c|cccccccccc|c}
			\toprule
			\textbf{Model} & \textbf{Instruction} & \textbf{Approach} & \textbf{Retreat} & \textbf{Pass} & \textbf{Land} & \textbf{Turn} & \textbf{Move} & \textbf{Shift} & \textbf{Rotate} & \textbf{Surround} & \textbf{A/D} & \textbf{Average} \\ 
            \midrule
			
			\multirow{2}{*}{Seq2Seq-UAV} 
            & Fixed & 0.00 & 0.00 & 15.00 & 0.00 & 0.00 & 0.00 & 0.00 & 0.00 & 0.00 & 0.00 & 1.50 \\
			& Open  & 0.00 & 0.00 & 2.50  & 0.00 & 0.00 & 0.00 & 0.00 & 0.00 & 0.00 & 0.00 & 0.25 \\ 
            \hline
			
			\multirow{2}{*}{CMA-UAV}     
            & Fixed & 0.00 & 91.67 & 25.00 & 0.00 & 0.00 & 0.00 & 0.00 & 0.00 & 0.00 & 0.00 & 11.67 \\
			& Open  & 0.00 & 91.67 & 25.00 & 0.00 & 0.00 & 0.00 & 0.00 & 0.00 & 0.00 & 0.00 & 11.67 \\ 
            \hline
			
			\multirow{2}{*}{Travel-UAV}  
            & Fixed & 35.71 & 100.00 & 25.00 & 53.70 & 20.00 & 13.33 & 85.71 & 46.67 & 58.33 & 84.21 & 52.27 \\
			& Open  & 42.86 & 100.00 & 32.50 & 40.74 & 6.67 & 6.67 & 75.51 & 40.00 & 100.00 & 78.95 & 52.39 \\ 
            \hline
			
			\multirow{2}{*}{OpenVLA-UAV} 
            & Fixed & 45.24 & 100.00 & 37.50 & 46.30 & 73.33 & 66.67 & 77.55 & 20.00 & 100.00 & 89.47 & \cellcolor[HTML]{FFCCC9}{65.61} \\
			& Open  & 47.62 & 100.00 & 55.00 & 40.74 & 73.33 & 60.00 & 85.71 & 6.67  & 100.00 & 84.21 & 65.33 \\ 
            \hline
			
			\multirow{2}{*}{$\pi_0$-0-UAV}    
            & Fixed & 59.52 & 100.00 & 50.00 & 66.67 & 33.33 & 60.00 & 18.37 & 46.67 & 58.33 & 26.32 & 51.92 \\
			& Open  & 71.43 & 100.00 & 62.50 & 72.22 & 80.00 & 80.00 & 40.82 & 60.00 & 75.00 & 15.79 & \cellcolor[HTML]{FFCCC9}{65.78} \\ 
            \hline
			
			\multirow{2}{*}{\shortstack{\textbf{WorldVLN}\\\textbf{(Ours)}}} 
            & Fixed 
            & 97.62 
            & 91.67 
            & 40.00 
            & 92.59 
            & 60.00 
            & 100.00 
            & 85.71 
            & 46.67 
            & 58.33 
            & 94.74 
            & \cellcolor[HTML]{FD6864}{\textbf{79.12}} \\
			& Open  
            & 95.24 
            & 91.67 
            & 40.00 
            & 98.15 
            & 53.33 
            & 100.00 
            & 85.71 
            & 33.33 
            & 41.67 
            & 94.74 
            & \cellcolor[HTML]{FD6864}{\textbf{78.02}} \\ 
            \bottomrule
		\end{tabular}
	}
\end{table*}

\begin{table*}[t]
    \centering
    \caption{Results on the IndoorUAV-VLA benchmark. We report Success Rate (\%) and NDTW (\%) on three difficulty splits, as well as the \textbf{Average} results on the full test set. See Appendix~\ref{subsec:indooruav_details}.}
    \label{tab:indooruav_vla}
    \resizebox{\linewidth}{!}{
        \begin{tabular}{l|cc|cc|cc|cc}
            \toprule
            \multirow{2}{*}{\textbf{Model}}
            & \multicolumn{2}{c|}{\textbf{Easy}} 
            & \multicolumn{2}{c|}{\textbf{Medium}} 
            & \multicolumn{2}{c|}{\textbf{Hard}} 
            & \multicolumn{2}{c}{\textbf{Average}} \\
            & \textbf{SR/\%$\uparrow$} & \textbf{NDTW/\%$\uparrow$}
            & \textbf{SR/\%$\uparrow$} & \textbf{NDTW/\%$\uparrow$}
            & \textbf{SR/\%$\uparrow$} & \textbf{NDTW/\%$\uparrow$}
            & \textbf{SR/\%$\uparrow$} & \textbf{NDTW/\%$\uparrow$} \\
            \midrule
            
            GPT-4o              
            & 30.30 & 12.30 
            & 4.00  & 6.57 
            & 1.96 & 4.84 
            & 11.69 & 9.20 \\
            
            Seq2Seq        
            & 1.60  & 2.63  
            & 1.20  & 2.74 
            & 1.03 & 3.03 
            & 1.33  & 2.74 \\
            
            CMA            
            & 1.28  & 1.75  
            & 0.75  & 1.94 
            & 1.03 & 2.01 
            & 0.99  & 1.88 \\
            
            OpenVLA         
            & 22.52 & 2.89  
            & 1.19  & 1.12 
            & 0.00 & 0.12 
            & 7.81  & 2.42 \\
            
            $\pi_0$-FAST  
            & 18.09 & 8.83  
            & 5.26  & 2.93 
            & 1.14 & 2.68 
            & 8.62  & 4.71 \\
            
            NaVid          
            & 25.31 & 13.10 
            & 18.21 & 3.21 
            & 2.31 & 1.72 
            & 15.82 & 5.28 \\
            
            $\pi_0$         
            & 46.58 
            & 14.52 
            & 21.64 
            & 7.64 
            & 7.55 
            & 4.27 
            & \cellcolor[HTML]{FFCCC9}{27.16} 
            & \cellcolor[HTML]{FFCCC9}{9.44} \\ 
            \hline
            
            \textbf{WorldVLN (Ours)} 
            & 49.43 
            & 13.04 
            & 37.72 
            & 14.52 
            & 41.19 
            & 12.80 
            & \cellcolor[HTML]{FD6864}{\textbf{41.76}} 
            & \cellcolor[HTML]{FD6864}{\textbf{13.48}} \\
            
            \bottomrule
        \end{tabular}
    }
\end{table*}

The quantitative results demonstrate the strong performance of WorldVLN across both outdoor and indoor UAV benchmarks, as listed in Table~\ref{tab:uavflow_results} and Table~\ref{tab:indooruav_vla}.
\begin{itemize}[leftmargin=*]
    \item \textbf{Strong performance across benchmarks.} WorldVLN achieves the best results on both UAV-Flow-Sim and IndoorUAV-VLA. On UAV-Flow-Sim, it reaches $79.12\%$ and $78.02\%$ average SR under fixed-template and open-vocabulary instructions, outperforming the strongest baselines by $13.51$ and $12.24$ percentage points, respectively. On IndoorUAV-VLA, it achieves $41.76\%$ full-set SR, improving over the best baseline by $14.60$ percentage points. These consistent gains suggest that the WAM paradigm adapts effectively to both outdoor and indoor UAV settings.
    \item \textbf{Advantages over VLA baselines.} WorldVLN consistently outperforms VLA-based models, e.g. initialized from OpenVLA or $\pi_0$. Compared with OpenVLA, WorldVLN improves average SR by $13.10$ percentage points on UAV-Flow-Sim and $33.95$ percentage points on IndoorUAV-VLA. Compared with $\pi_0$, it also improves by $19.72$ and $14.60$ points, respectively. This supports the benefit of prediction-based world-action modeling over direct observation-to-action mapping.
    \item \textbf{Larger gains on challenging cases.} The advantage is more evident on difficult settings. On IndoorUAV-VLA, WorldVLN improves SR by $16.08$ points on Medium and $33.64$ points on Hard over the best baselines. On UAV-Flow-Sim, it performs especially well on spatially precise tasks such as \textit{Approach}, \textit{Land}, \textit{Move}, \textit{Shift}, and \textit{Ascend/Descend}. These results indicate that predicting latent action consequences is particularly useful for complex aerial navigation.
\end{itemize}

\subsection{Case analysis}

\begin{figure}[h]
  \centering
  \includegraphics[width=\linewidth]{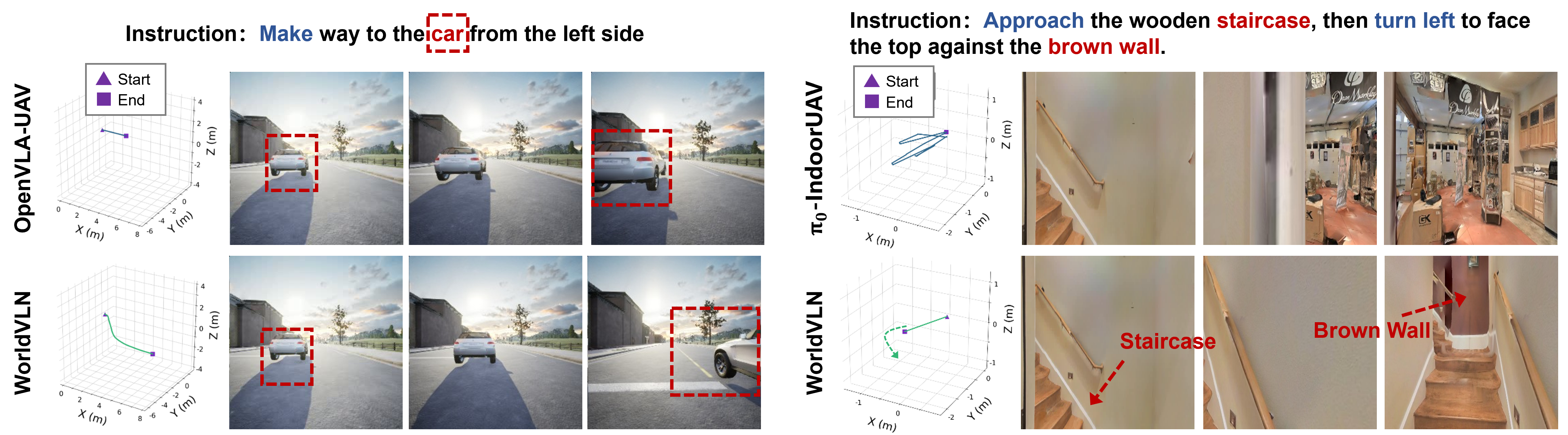}
  \caption{\textbf{Qualitative case analysis}. Compared with VLA baselines, WorldVLN shows stronger spatial grounding and more accurate waypoint actions in both outdoor object-centric maneuvers and indoor landmark navigation.}
  \label{fig:case_analysis}
\end{figure}

Figure~\ref{fig:case_analysis} qualitatively compares WorldVLN with representative VLA baselines in outdoor and indoor scenarios. In the outdoor case, the instruction requires the UAV to interact with the car. OpenVLA-UAV moves directly toward the vehicle and fails to execute a precise spatial maneuver around it. In contrast, WorldVLN correctly grounds the car as the target landmark and generates a smoother trajectory with accurate relative positioning.
In the indoor case, the instruction requires the agent to approach the staircase and turn left toward the brown wall. The $\pi_0$-IndoorUAV baseline fails to maintain the intended spatial relation with the staircase and wall. WorldVLN consistently identifies the relevant landmarks, approaches the staircase, and performs the left-turn behavior in accordance with the instructed spatial layout. These cases suggest that latent world-action prediction enables more accurate spatial grounding and waypoint generation than direct VLA-style action mapping.

\subsection{Ablation study}

\begin{figure}[t]
  \centering
  \includegraphics[width=\linewidth]{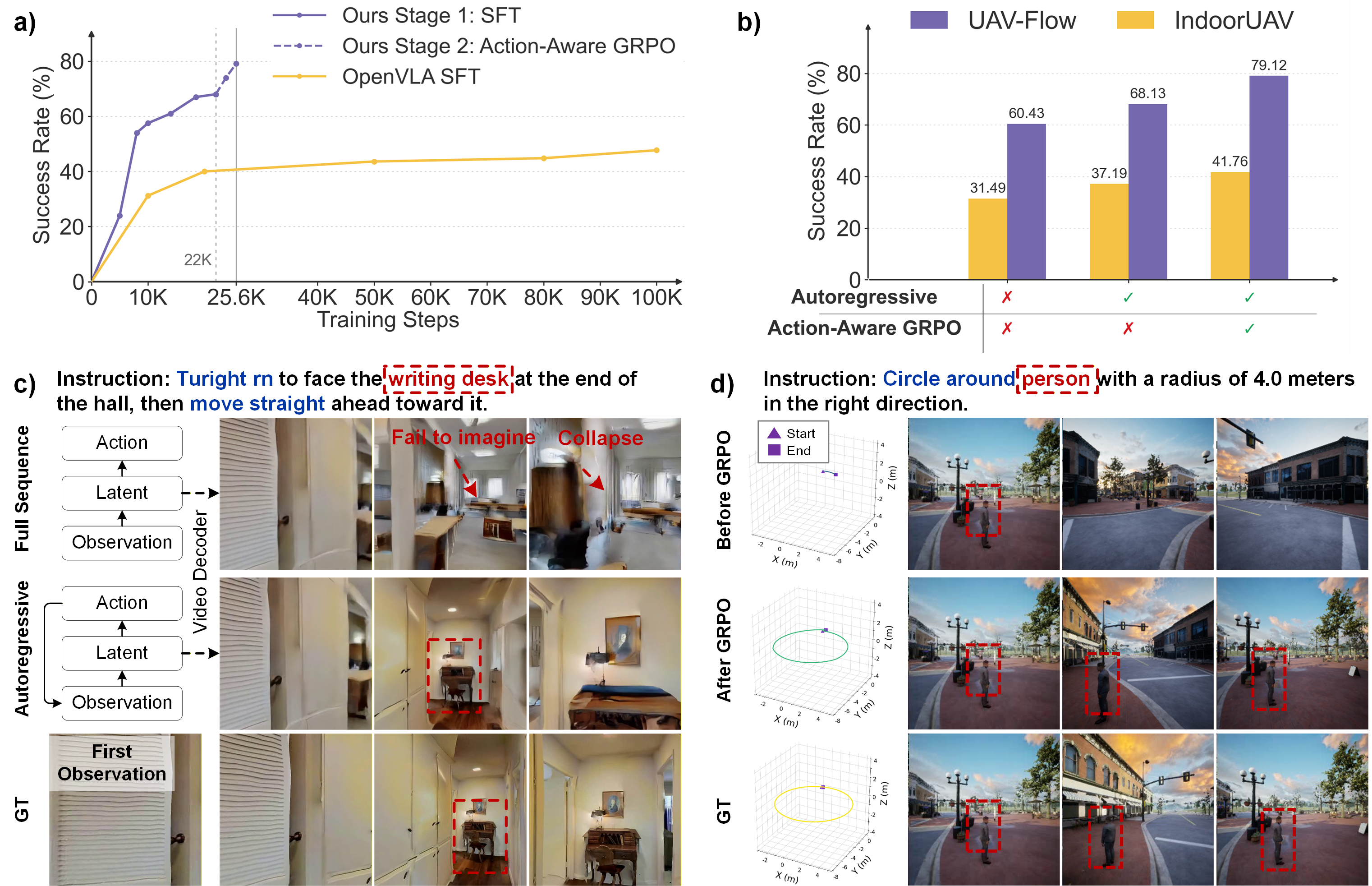}
  \caption{\textbf{Ablation studies}. a) Training dynamics compared with OpenVLA on UAV-Flow. b) Quantitative effects of autoregressive modeling and Action-aware GRPO on UAV-Flow and IndoorUAV. c) Latent prediction probe: autoregressive updating preserves more coherent visual-spatial representations than full-sequence prediction. d) Action-aware GRPO improves spatial action accuracy, producing a trajectory closer to the intended circular maneuver.}

  \label{fig:ablation}
\end{figure}

\paragraph{Does WAM learn more efficiently than VLA?}
We train OpenVLA from scratch on UAV-Flow and compare its training dynamics with WorldVLN, as shown in Figure~\ref{fig:ablation}(a). WorldVLN after Stage-1 supervised training reaches higher success rates under the same step budget than OpenVLA-SFT. This suggests that the WAM formulation provides a more effective learning structure for aerial VLN than direct VLA-style observation-to-action mapping.

\paragraph{Why is autoregressive prediction necessary?}
To isolate the effect of autoregressive modeling, we use the same backbone and action decoder, and compare full-sequence SFT with autoregressive SFT. Figure~\ref{fig:ablation}(b) quantitatively shows that autoregressive world-action modeling improves success rates on both UAV-Flow and IndoorUAV by $5.7+$ percentage points. To probe the learned latent representations, Figure~\ref{fig:ablation}(c) decodes predicted latents into visual observations for visualization only. The full-sequence variant exhibits semantic drift and scene collapse, indicating unstable long-horizon latent prediction. In contrast, the autoregressive variant repeatedly incorporates newly observed states and preserves more coherent visual-spatial representations, including the instruction-relevant landmark. This suggests that closed-loop autoregressive updating improves latent world prediction and provides more reliable representations for action decoding.

\paragraph{What does Action-aware GRPO learn?}
To evaluate the effect of action-aware GRPO, we compare the model trained only with Stage-1 supervised training and the model trained with the full two-stage framework. Figure~\ref{fig:ablation}(b) shows that adding Action-aware GRPO further boosts performance on both benchmarks. This is also reflected in Figure~\ref{fig:ablation}(a), where Action-aware GRPO yields an additional gain of over $10$ points after the Stage-1 SFT performance has nearly saturated. Moreover, Figure~\ref{fig:ablation}(d) visualizes navigation behavior before and after RL. Before Action-aware GRPO, the model fails to execute a geometrically accurate circular trajectory. After RL, the model produces a trajectory that better follows the intended ``circle around'' behavior and more closely matches the ground-truth path. This indicates that Action-aware GRPO teaches the model to optimize action consequences beyond visual plausibility, improving action accuracy and goal-directed behavior.

\subsection{Zero-shot generalization in real-world deployment}

\begin{figure}[t]
  \centering
  \includegraphics[width=\linewidth]{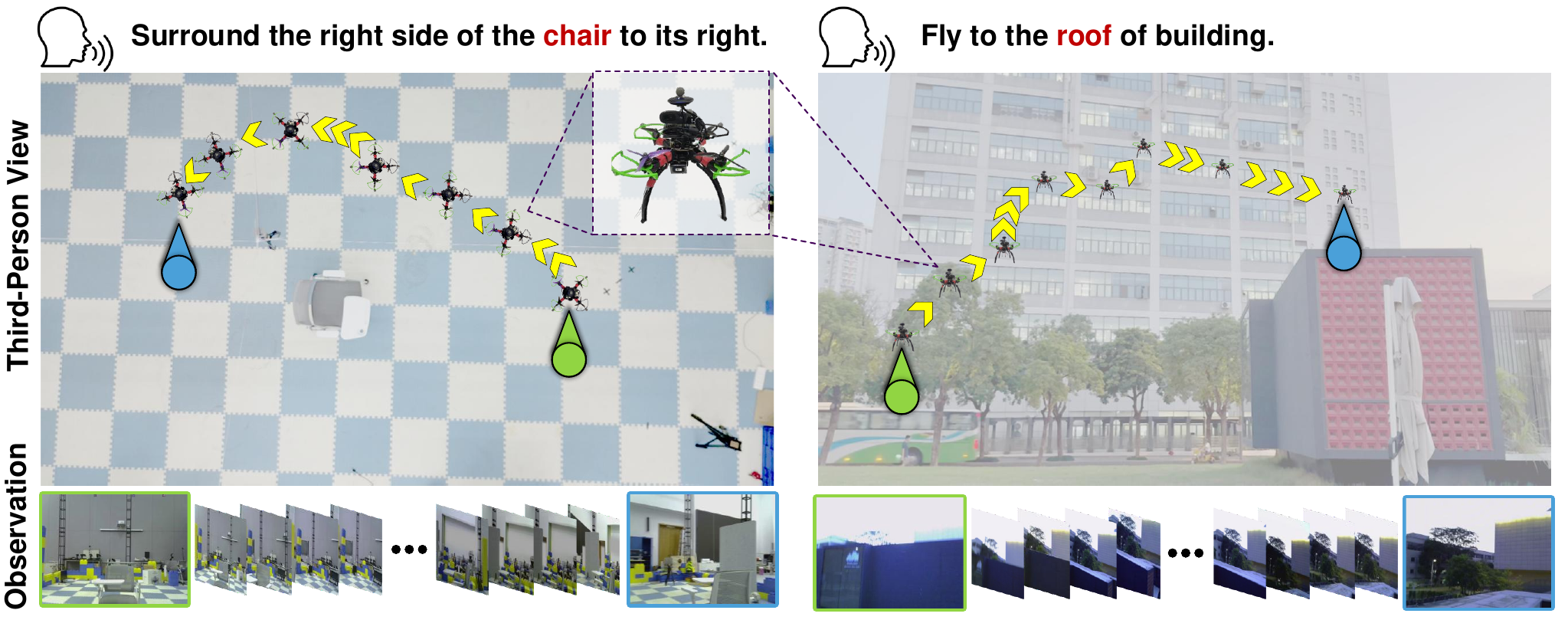}
  \caption{\textbf{Real-world UAV deployment}. WorldVLN is trained only in simulation and tested zero-shot on a real drone in both indoor and outdoor scenarios.}
  \label{fig:real_drone}
\end{figure}

To evaluate WorldVLN in real-world environments, we deploy it on a self-built quadrotor with a 250 mm wheelbase, equipped with a Logi C270 RGB camera, a Jetson Orin NX 16GB onboard computer, and a CUAV PX4 flight controller. The WorldVLN policy runs on a remote server: RGB observations are transmitted from the UAV to the server, and predicted waypoint actions are sent back for execution. We conduct indoor tests in a $10\,\mathrm{m}\times15\,\mathrm{m}\times3\,\mathrm{m}$ arena with a 14-camera MoCap system, and outdoor tests in an open area using GPS with a TFmini-S LiDAR for altitude estimation.

Figure~\ref{fig:real_drone} shows two representative real-world cases. Although WorldVLN is trained only with simulator data, it can follow language instructions and generate executable waypoint actions on the real UAV platform. The indoor case requires the UAV to approach and align with a target object in a confined room. This setting is challenging because the agent must rely on close-range visual landmarks and avoid large viewpoint deviations. The outdoor case further investigates the model's ability to navigate in the vertical direction. These results provide evidence that the learned world-action representation can transfer from simulation to real-world UAV deployment even without additional real-world fine-tuning. See Appendix~\ref{sec:real_uav} for details.

\section{Conclusions, Limitations, and Future Works}

We present WorldVLN, the first autoregressive world action model for aerial vision-language navigation. We introduce a two-stage training framework that first grounds the video prior in instruction-conditioned navigation dynamics and then develops Action-aware GRPO to align the model with action-level navigation outcomes. Experiments on indoor and outdoor benchmarks demonstrate \textbf{strong and transferable} performance, with over $12$ percentage-point gains over VLA baselines under less training-step budgets and larger advantages on hard tasks. Real-world UAV deployment further provides promising evidence of zero-shot transfer. Together, WorldVLN provides a concise implicit-prediction architecture and an Action-aware GRPO training strategy, offering a promising route for spatial action tasks and potentially broader embodied domains such as robotic manipulation.

\textbf{Limitations.}
Our experiments are validated on relatively short-range aerial navigation, while long-horizon VLN remains to be further explored. Real-world deployment also relies on server-side inference due to the computational cost of the backbone, limiting fully onboard execution.

\textbf{Future works.}
We will explore more scalable architectures for long-horizon latent prediction, as well as model compression and inference acceleration for fully onboard UAV deployment.

\bibliographystyle{plainnat}
\bibliography{ref}

\appendix

\input{appendix}



\end{document}

%% file: appendix.tex
\clearpage
\section{Appendix}

\subsection{Broader Impacts and Responsible Deployment}
\label{app:broader_impacts}

WorldVLN may benefit UAV-based embodied navigation applications such as infrastructure inspection, search and rescue, and disaster assessment. In these scenarios, language-conditioned aerial agents can reduce the cost of manual operation and decrease the need for humans to enter hazardous or hard-to-access environments. However, autonomous UAV navigation may also introduce potential risks, including physical safety concerns, privacy violations, unauthorized surveillance, and misuse in restricted or safety-sensitive areas. Therefore, the results in this paper should not be interpreted as evidence that the system is ready for unsupervised deployment in public or safety-critical environments.

Our real-world experiments are conducted only in controlled indoor arenas or relatively enclosed outdoor areas. Low-level flight stabilization is handled by the PX4 flight controller, while high-level model inference and monitoring are performed through a ground-station server. Practical deployment should include human supervision, geofencing, speed and altitude limits, emergency stop mechanisms, and reliable state-estimation checks, and should comply with local UAV regulations. For code and model release, we recommend that they be used primarily for research and simulator evaluation; real-world UAV deployment interfaces should only be used after sufficient safety validation and under appropriate hardware supervision.

\subsection{Limitations}
\label{app:limitations}

Although WorldVLN achieves strong results on both indoor and outdoor UAV benchmarks, the current study mainly focuses on short-range aerial navigation and short-horizon waypoint generation. Therefore, the capability of the model remains to be further validated in long-horizon VLN scenarios, multi-stage complex instructions, large-scale outdoor exploration, and long-term closed-loop decision-making.

In addition, WorldVLN is mainly trained on simulator data and public benchmark trajectories, and the real-world evaluation is conducted only in controlled indoor arenas and relatively enclosed outdoor areas. Its robustness under more challenging real-world conditions, such as strong illumination changes, adverse weather, dynamic obstacles, crowded spaces, or GPS-denied environments, has not been fully evaluated. Due to the large scale of the autoregressive world backbone, the current real-world deployment still relies on server-side inference and cannot yet run fully onboard. Moreover, we do not conduct extensive multi-seed experiments in this work. Future work should further improve model compression, inference acceleration, safety constraints, and statistical validation.

\subsection{Details of model architecture}
\label{app:model_architecture}
\begin{figure}[H]
    \centering
    \includegraphics[width=0.98\linewidth]{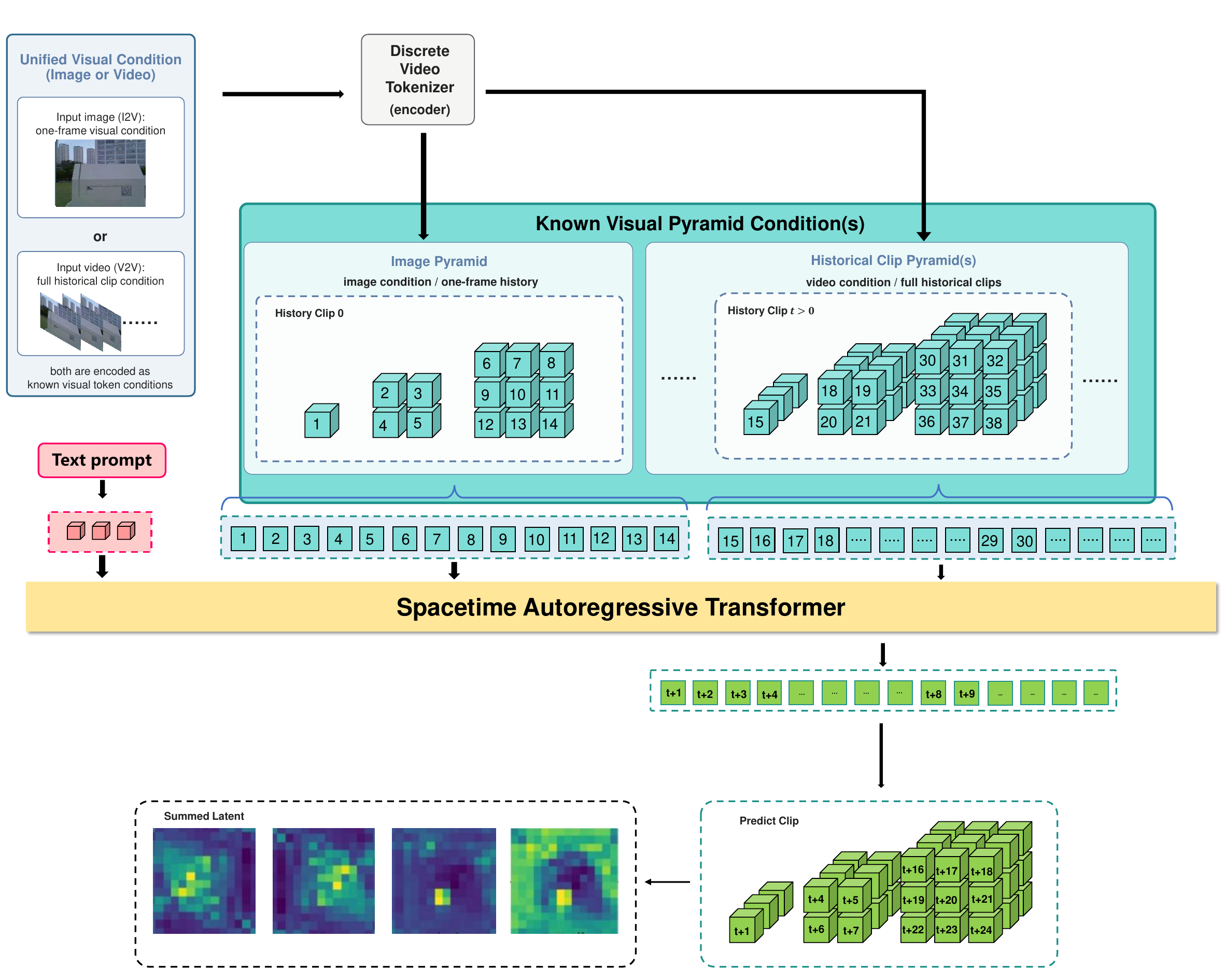}
    \caption{Architecture of the latent-space spatiotemporal autoregressive world backbone. The input image or historical video is encoded into known visual pyramid conditions, which are used together with text tokens to predict future target clip pyramids. The predicted token blocks are aggregated into output latent representations for subsequent action decoding.}
    \label{fig:world_backbone_arch}
\end{figure}

WorldVLN adopts a latent-space spatiotemporal autoregressive architecture as its world-model backbone. Following the InfinityStar~\cite{liu2025infinitystar} and the WAN VAE~\cite{wan2025wan}, the overall architecture consists of a text encoder, a discrete video tokenizer, a spatiotemporal autoregressive Transformer, and an action decoder. As shown in Figure~\ref{fig:world_backbone_arch}, given a language instruction and egocentric visual observations, the text encoder converts the instruction into text tokens, while the visual observations are converted by the video tokenizer into known visual pyramid conditions. The spatiotemporal autoregressive Transformer then predicts the token blocks of future target clips conditioned on both the text tokens and the known visual token conditions. These predicted token blocks are further aggregated into future world-state latent representations, which are used as the input to the action decoder for action generation.

The visual tokenizer uses the video VAE encoder from the adopted pretrained tokenizer as the continuous visual compression module. Specifically, the input image or video is first encoded into a compact latent representation. Multi-scale residual quantization is then performed on this latent representation to obtain a set of discrete residual token blocks. For a single-frame image condition, these token blocks are organized as an image pyramid; for multi-frame historical video conditions, they are organized as historical clip pyramids. This representation unifies visual observations into multi-scale token conditions, enabling both image inputs and historical video inputs to be processed by the same spatiotemporal autoregressive Transformer.

The spatiotemporal autoregressive Transformer predicts future target clip pyramids from the known visual pyramid conditions. Within each target clip, the model follows a coarse-to-fine scale order for next-scale prediction: it first predicts low-resolution token blocks that mainly capture global structure, and then progressively predicts higher-resolution token blocks that provide local details. Along the temporal dimension, the model performs clip-order autoregression: the first target clip is predicted from the known visual conditions, and subsequent target clips are predicted conditioned on preceding target clips. Therefore, this architecture jointly models the spatial scale dependency within each clip and the temporal evolution across clips. After prediction, the multi-scale token blocks of each target clip are merged into the corresponding latent representation. This latent representation is not decoded as the final RGB video output; instead, it is treated as a future world-state representation and fed into the action decoder to generate low-level navigation actions.

\begin{figure}[t]
    \centering
    \includegraphics[width=0.98\linewidth]{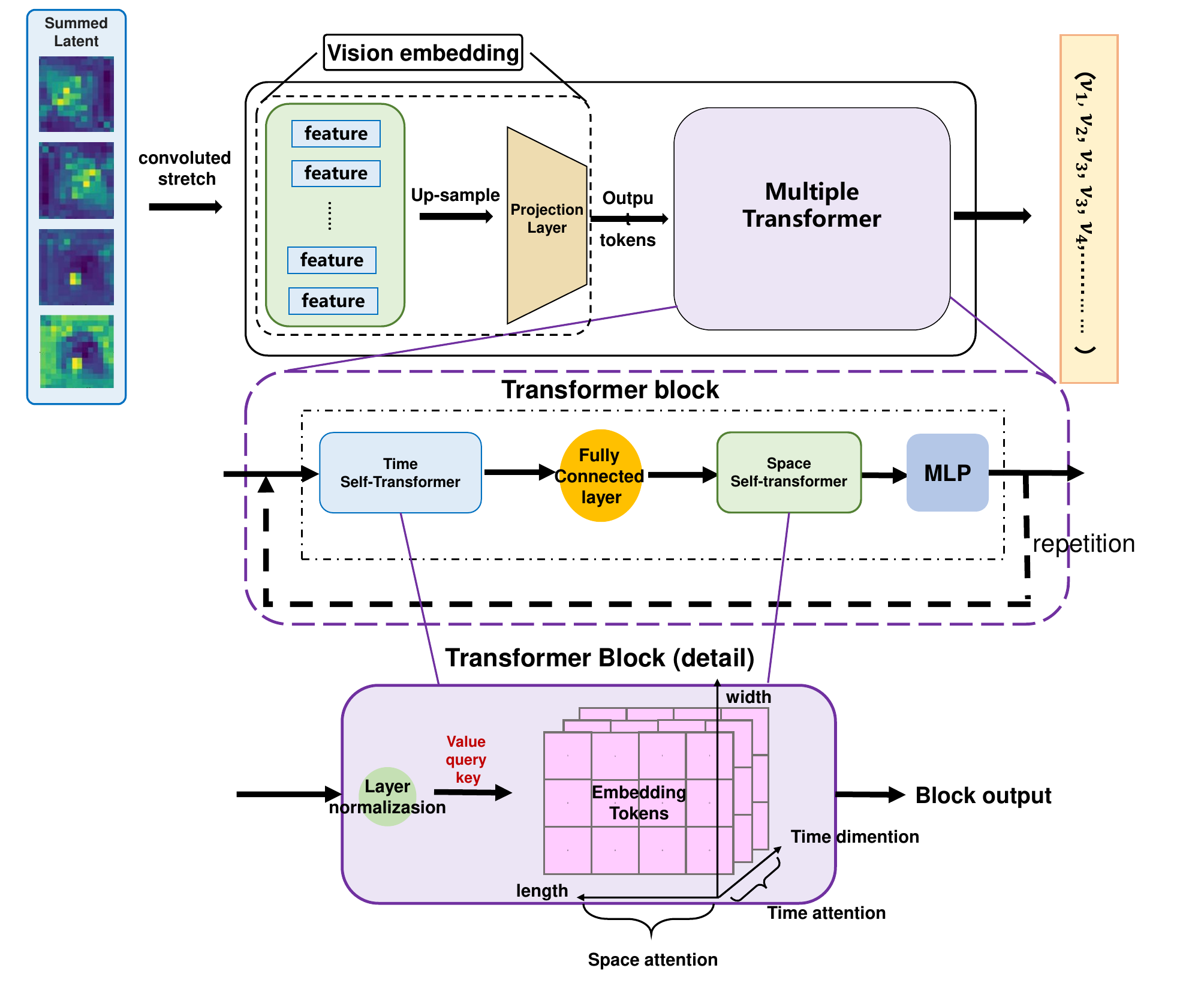}
    \caption{Architecture of the action decoder. The world-model output latent is first converted into spatiotemporal embedding tokens by the vision embedding module. Multiple Transformer blocks with factorized temporal and spatial attention then model action-related spatiotemporal features, which are finally regressed into continuous UAV navigation actions.}
    \label{fig:action_decoder_arch}
\end{figure}

The action decoder receives the future latent representation produced by the world backbone and converts it into executable low-level navigation actions. As shown in Figure~\ref{fig:action_decoder_arch}, the module takes the world-model output latent as input and regards it as a compact spatiotemporal representation that contains short-horizon future state changes. This latent representation encodes spatiotemporal information related to viewpoint changes, spatial-structure changes, and motion trends within the prediction horizon, and thus provides a direct basis for action inference. The action decoder finally outputs a continuous action vector, where each action corresponds to the relative 3D displacement and relative yaw change of the UAV.

Structurally, the action decoder consists of a vision embedding module, a spatiotemporal Transformer backbone, and an action regression head. The vision embedding module first reorganizes the input segment-level latent through feature reshaping, convolutional mapping, upsampling, and projection, converting it into unified spatiotemporal embedding tokens while preserving both the temporal order and spatial layout of the latent representation. Then, multiple Transformer blocks perform action-related feature modeling with factorized spatiotemporal attention: temporal attention captures motion evolution and viewpoint changes across latent frames, while spatial attention models the geometric structure and spatial relations within each latent frame. Finally, the aggregated spatiotemporal representation is regressed by an MLP action head into continuous action vectors, enabling the action decoder to infer the UAV navigation actions directly from the future latent state changes predicted by the world model.

\subsection{Details of training framework}
\label{subsec:training_framework}

\paragraph{Supervised fine-tuning of the World Backbone.} WorldVLN adopts a spacetime autoregressive video prediction setup during supervised fine-tuning. Given a navigation instruction and its corresponding egocentric navigation video, we divide the video into one initial segment and multiple future segments. The initial segment contains only the frame 0 and provides the initial observation condition, while each future segment contains $K=16$ frames and represents a short-horizon future observation segment. The temporal compression ratio of the video tokenizer is set to 4; therefore, each 16-frame clip corresponds to 4 temporal steps in the latent/token space. Each training sample contains 49 frames in total, consisting of one initial frame and three consecutive 16-frame future segments. This configuration aligns the temporal granularity of the world model with the short-horizon decision window required for navigation.

When training the autoregressive world backbone, we adopt segment-level surpervised fine-tuning. For each target future segment, the model is conditioned on the language instruction and the complete ground-truth observation history preceding that segment, and predicts the discrete multi-scale token representation of the current clip. The model does not use its own previously generated clips as historical context during this stage. We retain the video decoder as a training-time visual supervision interface, so that the backbone is optimized toward future visual predictions that remain decodable by the video decoder. Through this first-stage SFT, the model learns to predict plausible short-horizon future visual changes conditioned on the instruction and real historical observations, thereby forming a latent imagination capability for navigation. During inference, the video decoder is not used for action generation; instead, the predicted latent world transition is directly fed into the subsequent action decoder.

\paragraph{Supervised training of the Action Decoder.}
The video-to-action teacher follows the TSformer-VO-style visual odometry/action decoding backbone, while the latent projection is initialized with priors from the adopted video decoder. The training of the Action Decoder starts from a video-to-action teacher model. We first train this teacher model using real video clips paired with expert action sequences, enabling it to learn the mapping from continuous visual observations to navigation actions. The teacher model then provides action-aware spatiotemporal representation supervision for the latent-space Action Decoder, which alleviates the representation alignment difficulty caused by directly learning action prediction from compressed latents.

Based on this teacher model, we transfer the action decoding process into the latent space of the world model. Specifically, each real video clip is first encoded by the frozen video encoder to obtain the ground-truth video latent. We then use the embedding/token representation produced by the pretrained video-to-action teacher model as the distillation target, and align the Vision Embedding module of the Action Decoder to perform latent-to-token representation mapping, so that its output can be compatible with the subsequent action decoding backbone. After this representation alignment, we train the entire Action Decoder with ground-truth video latents and their corresponding expert action sequences, ultimately establishing the mapping from latent world representations to continuous navigation actions.

\paragraph{Action-aware GRPO.}
We provide the reward implementation details used in Action-aware GRPO. The formulation follows the segment-wise notation in the main text and is applied to both UAV-Flow and IndoorUAV-VLA, with benchmark-specific action formats and success thresholds. For each navigation case, the current policy samples a group of $G=4$ online autoregressive rollouts. The $i$-th rollout contains multiple decision segments, and the $j$-th action segment is denoted by $a^{(i)}_{(j-1)K:jK-1}$. Each rollout starts from the given initial pose and observation. At each segment, the model predicts a short-horizon latent world transition, decodes it into waypoint actions, executes the actions in the environment, receives the next observation segment, and updates the autoregressive context. Future frames are not provided during rollout.

We use three reward terms: an trajectory-consistency reward, a task-progress reward, and a CE-style reference alignment reward. 
In our experiments, we set the weights of the three reward terms as:
\begin{equation}
    \lambda_{\mathrm{traj}}=0.2,
    \qquad
    \lambda_{\mathrm{task}}=0.7,
    \qquad
    \lambda_{\mathrm{ref}}=0.1.
\end{equation}

\textbf{Trajectory reward.}
For each segment, we first compute the action error between the sampled action segment and the expert action segment. The trajectory distance is defined as
\begin{equation}
    d^{(i)}_{\mathrm{traj},j}
    =
    0.45\,\mathrm{MSE}^{(i)}_{xyz,j}
    +
    0.45\,\mathrm{MSE}^{(i)}_{\mathrm{yaw},j}
    +
    0.1\,\mathrm{MSE}^{(i)}_{\mathrm{all},j},
\end{equation}
where $\mathrm{MSE}_{xyz}$ measures the translation error, $\mathrm{MSE}_{\mathrm{yaw}}$ measures the yaw error, and $\mathrm{MSE}_{\mathrm{all}}$ is computed over the full normalized action representation. Equivalently, $d^{(i)}_{\mathrm{traj},j}$ serves as the implementation of the trajectory distance used in Eq.~\ref{eq:trajectory_reward}.
This term encourages the sampled waypoint segment to remain geometrically consistent with the expert trajectory. We assign a relatively large weight to yaw error because yaw changes directly determine the agent's egocentric field of view and strongly affect subsequent observations, while yaw control is empirically harder to learn than translational displacement.

\textbf{Task reward.}
The task reward measures whether the rollout successfully completes the intended navigation goal. We determine task progress mainly by comparing the distance between the rollout endpoint and the target endpoint, starting from the same initial pose. This endpoint-based criterion is applicable to both UAV-Flow and IndoorUAV-VLA, while the specific success thresholds follow the corresponding benchmark protocol.
Specifically, we can use both quantitative and qualitative signals to measure the task reward. The quantitative signal is a dense geometric score based on the Euclidean endpoint distance. The qualitative signal is a binary success indicator determined by the benchmark-specific success rule.

\textbf{Reference reward.}
To regularize the updated policy toward the reference behavior, we use a CE-style alignment cost computed from the reference-policy log-probability. We introduce this term because optimizing only the trajectory and task rewards can overly shift the autoregressive backbone away from its original video-generation prior. Empirically, when the fine-tuned backbone is connected back to the video decoder for visualization, the generated frames may show degraded visual consistency or even collapse, indicating that the latent imagination capability has been weakened. The CE-style reference alignment reward mitigates this issue by constraining the updated policy to stay close to the reference behavior, thereby preserving the world-model prior learned during supervised training. Smaller CE costs indicate stronger agreement with the reference policy.

\textbf{Temporal decay and trajectory-level gate.}
We apply temporal decay $\gamma=0.9$ to the segment rewards.
This decay gives earlier decisions larger weights because errors in early segments influence later observations, actions, and accumulated trajectory drift. 

In total, these design provides dense credit assignment for the VLN tasks. The main optimization settings are a maximum of 3000 training iterations, learning rate $8\times10^{-7}$, KL coefficient $0.9$, PPO ratio clipping threshold $0.02$, gradient clipping $0.5$, video batch size 1, and maximum token length 20480. To make GRPO fine-tuning memory-efficient for the 8B world backbone, we use partial freezing with the first five model chunks frozen.

\subsection{Details of experimental setup}
\label{subsec:exp_setup}

To systematically evaluate the effectiveness of WorldVLN on UAV vision-language action generation, we conduct experiments on two complementary benchmarks: UAV-Flow and IndoorUAV-VLA. UAV-Flow follows the Flying-on-a-Word task formulation, where the model generates low-level flight actions conditioned on egocentric observations, UAV states, and atomic language instructions. It further evaluates the model's ability to capture flight intent, visual spatial relations, and dynamically feasible flight trajectories under language-conditioned UAV control. IndoorUAV-VLA is the indoor VLA subset of IndoorUAV, constructed by segmenting long-horizon indoor navigation trajectories into short sub-trajectories. Each instruction typically corresponds to 1--3 local UAV actions, which enables evaluation of local spatial understanding, orientation control, and fine-grained action generation in continuous 3D indoor environments. Together, these two benchmarks provide complementary evaluation settings for vision-language navigation.

\begin{figure}[H]
    \centering
    \includegraphics[width=\linewidth]{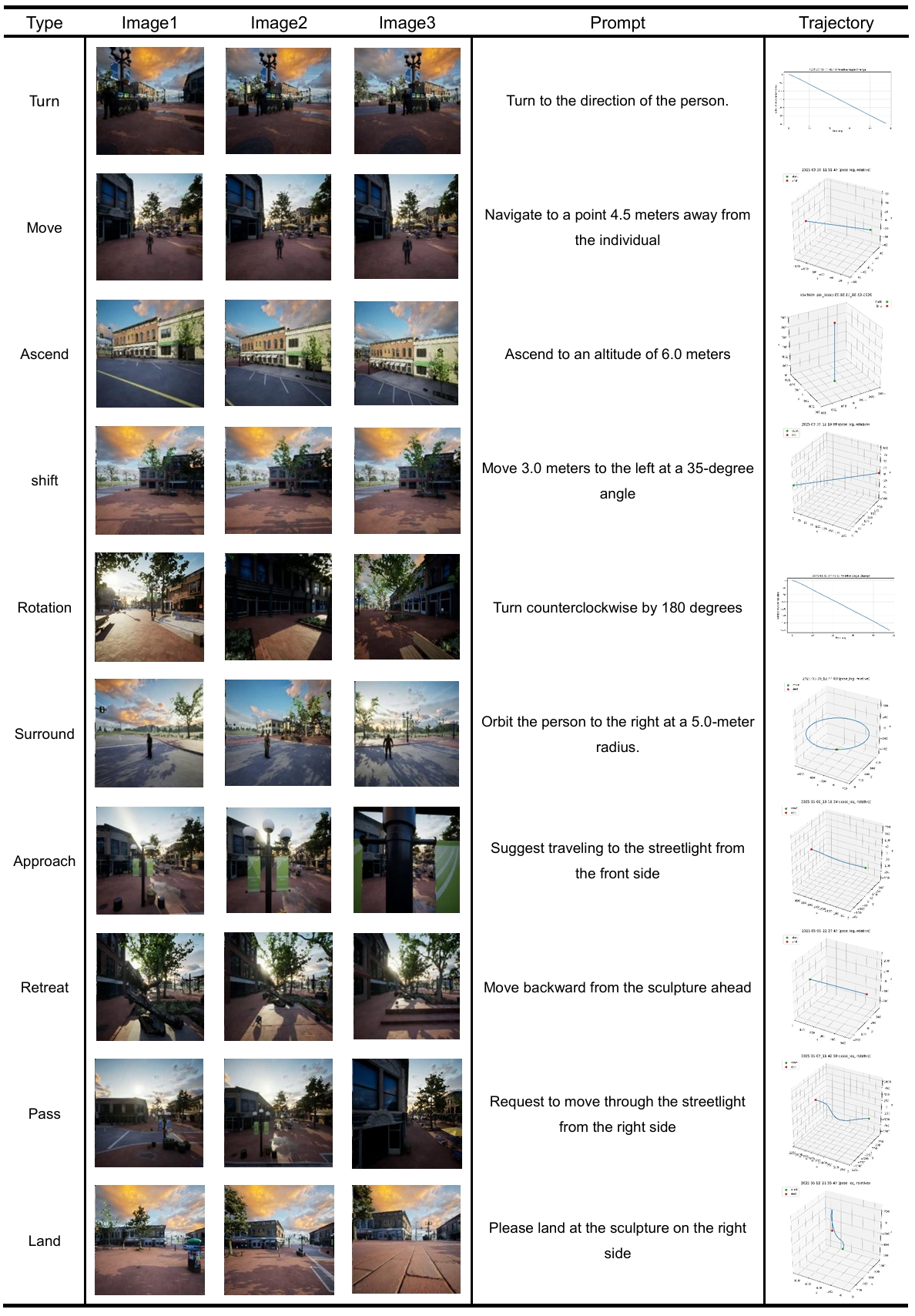}
    \caption{
    Qualitative examples from the UAV-Flow benchmark. The benchmark covers diverse fine-grained UAV flight actions, including target-oriented motion, primitive translation, vertical control, and object-relative navigation.
    }
    \label{fig:uavflow_cases}
\end{figure}

\begin{figure}[H]
    \centering
    \includegraphics[width=\linewidth]{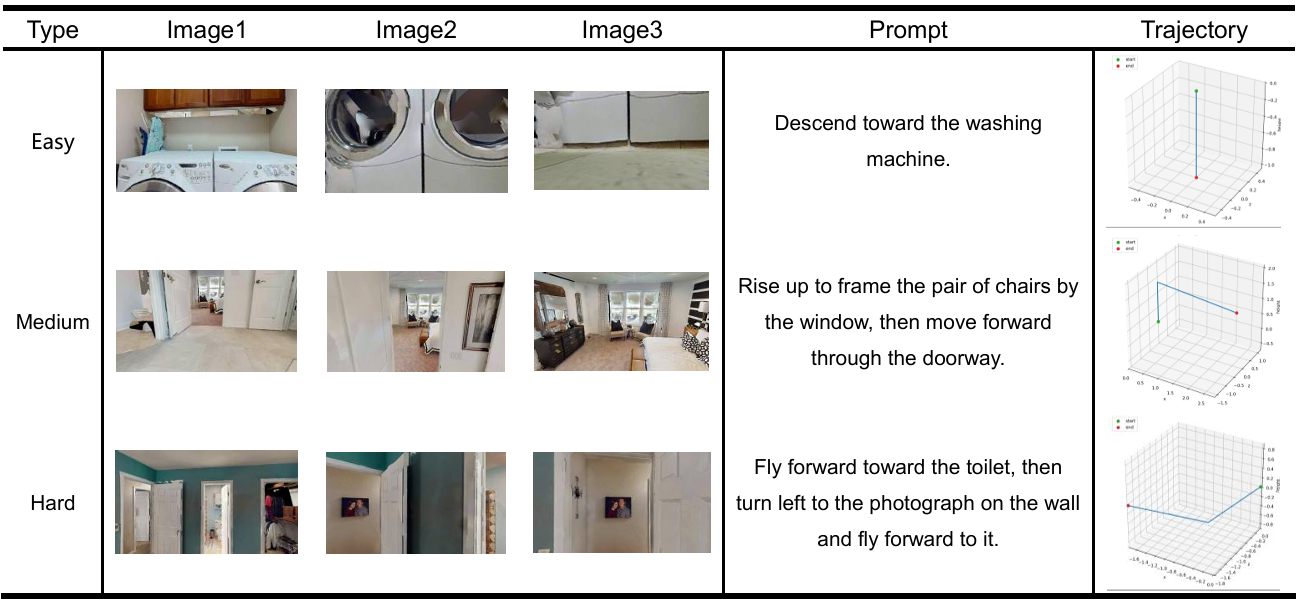}
    \caption{
    Qualitative examples from the IndoorUAV-VLA benchmark. Easy, Medium, and Hard correspond to increasing action-composition complexity, where the UAV needs to execute one, two, or three types of low-level actions.
    }
    \label{fig:indooruav_cases}
\end{figure}

For UAV-Flow, the evaluation covers both fixed-template and open-vocabulary instructions, and reports success rate (SR) across multiple fine-grained flight skill categories. As shown in Figure~\ref{fig:uavflow_cases}, the benchmark contains diverse flight skills such as approaching, landing, moving, shifting, and ascending/descending. Such category-wise evaluation provides a more detailed assessment of model performance under different motion semantics, spatial interaction patterns, and language variations. For IndoorUAV-VLA, we follow the original benchmark protocol and report SR and NDTW. As shown in Figure~\ref{fig:indooruav_cases}, the Easy, Medium, and Hard splits correspond to different levels of action-composition complexity. Its NDTW metric considers both 3D positional trajectories and yaw-angle changes, which makes it better suited for evaluating 4-DoF indoor UAV control involving forward motion, lateral motion, vertical movement, and yaw rotation. 

For both benchmarks, we compare WorldVLN with representative baselines covering different technical paradigms, including traditional VLN methods, UAV-specific or waypoint-based policies, and general VLA models. All methods are evaluated under the corresponding input-output format and evaluation protocol of each benchmark, ensuring fair comparison and enabling a comprehensive analysis of WorldVLN's world-action modeling capability.

\paragraph{Existing assets and licenses.}
Our experiments use public benchmarks, pretrained models, and open-source software under their original protocols and licenses. Specifically, UAV-Flow and IndoorUAV-VLA are used for training and evaluation; InfinityStar-8B and the video VAE serve as pretrained backbone/tokenizer components; and PX4, MAVLink, and VRPN support real-world flight control, communication, and external pose integration. We cite the corresponding papers or project pages and follow their terms of use.

\subsection{Details of results on UAV-Flow}

\label{subsec:uavflow_details}

UAV-Flow evaluates fine-grained language-conditioned UAV control under both fixed-template and open-vocabulary instruction settings. Our model achieves 79.12\% average SR under the Fixed setting and 78.02\% average SR under the Open setting, with only a small performance gap between the two settings, demonstrating that the model can robustly map different language expressions to consistent low-level flight behaviors. Specifically, we improve the performance on \textbf{Approach} to 97.62\% Fixed SR and 95.24\% Open SR, showing that the model can effectively couple target localization, distance estimation, and target-proximity control. We improve the performance on \textbf{Land} to 92.59\% Fixed SR and 98.15\% Open SR, indicating that the model can accurately control both horizontal target alignment and vertical descent, which verifies its strong 3D terminal-state control ability. We further achieve 100.00\% SR on \textbf{Move}, 85.71\% SR on \textbf{Shift}, and 94.74\% SR on \textbf{A/D} under both instruction settings, demonstrating reliable primitive translation, egocentric lateral control, and vertical degree-of-freedom control. These results show that our model is particularly effective in aligning language instructions, visual targets, and low-level UAV motion, especially for target proximity, precise landing, basic translation, sideward movement, and height control.

\subsection{Details of results on IndoorUAV}

\label{subsec:indooruav_details}

IndoorUAV-VLA evaluates short-horizon indoor UAV control using SR and NDTW, where SR measures whether the UAV successfully reaches the target pose and NDTW measures trajectory-level consistency with the reference path. Our model achieves 41.76\% SR and 13.48\% NDTW on the Full split, showing that it improves both final-pose success and trajectory quality. More importantly, we achieve 37.72\% SR and 14.52\% NDTW on the Medium split, indicating that the model can better compose multiple low-level actions within a short trajectory rather than only executing a single primitive motion. On the Hard split, our model further achieves 41.19\% SR and 12.80\% NDTW, demonstrating stronger multi-step coordination when translation, rotation, vertical movement, and other motion primitives must be jointly executed. These improvements show that the main advantage of our method lies in compositional short-horizon UAV control, where online autoregressive GRPO rollouts and closed-loop world-action updating help the model maintain state consistency and correct intermediate decisions across multiple action steps.
\subsection{Details of VLA vs WAM}

To clarify the experimental setup for comparing VLA and WAM, we use OpenVLA as a representative VLA baseline and WorldVLN as an autoregressive world action model. Both models are trained and evaluated on exactly the same training and test splits of UAV-Flow-Sim, and use the same waypoint/action normalization. The SR evaluation protocol also follows the original benchmark setting. This setup controls for factors such as data splits, action normalization, and evaluation criteria, allowing the comparison to focus more directly on the difference between the two modeling paradigms: OpenVLA follows a direct observation-to-action formulation, whereas WorldVLN first performs latent world prediction and then generates actions through an action decoder.

In terms of configuration, we follows the setting in the UAV-Flow benchmark for the OpenVLA training. Both models are trained with the AdamW optimizer and bf16 precision. The learning rate of WorldVLN is set to $1 \times 10^{-5}$, with a maximum training budget of 26K steps. The learning rate of OpenVLA is set to $5 \times 10^{-4}$, with a maximum training budget of 100K steps. For batch size, both models are trained on 8 GPUs. OpenVLA uses a per-GPU batch size of 1, resulting in a fixed global batch size of 8. WorldVLN uses token-budget-based sequence packing; under the current 49-frame configuration, its effective global batch size is approximately 8 clips per step. Overall, this setup preserves the native training configuration of each model family under the same data and evaluation protocol, enabling a comparison between VLA and WAM on UAV-Flow-Sim.

\subsection{Details of Real-World Deployment}\label{sec:real_uav}

\begin{figure}[H]
    \centering
    \includegraphics[width=\linewidth]{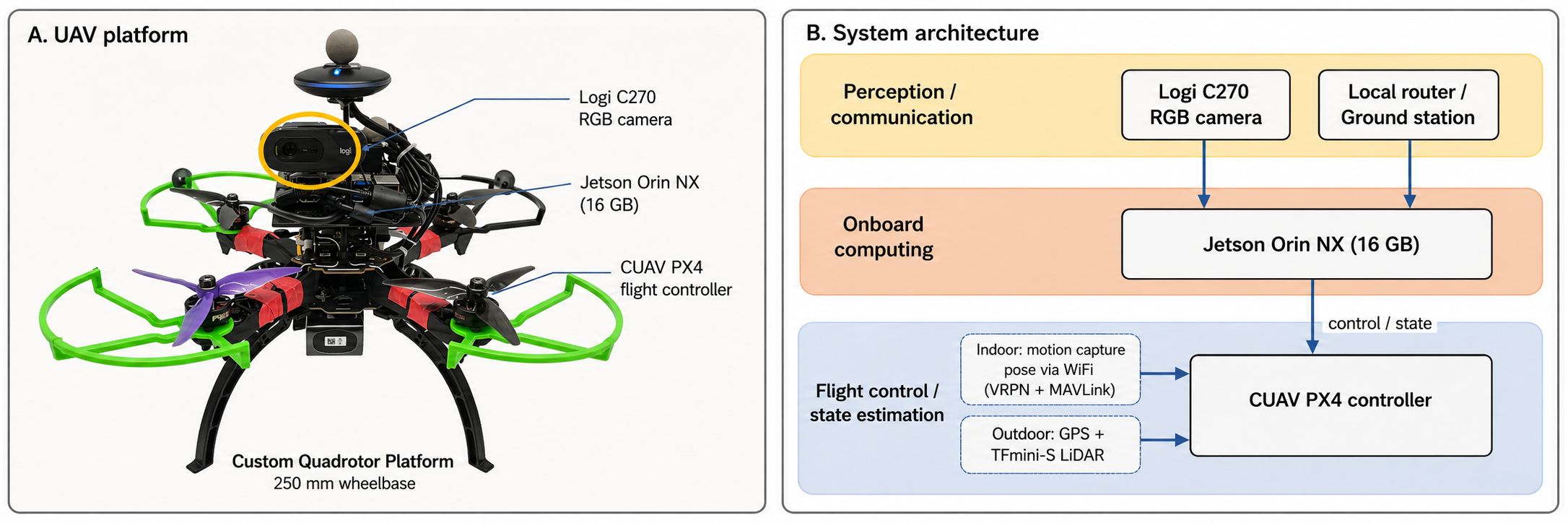}
    \caption{Real-world UAV platform and system architecture.}
    \label{fig:uavhardware}
\end{figure}
To evaluate the deployability of our proposed method in real-world environments, we develop a custom quadrotor platform with a 250~mm wheelbase. As shown in Figure~\ref{fig:uavhardware}, the platform is equipped with a Logi C270 RGB camera for egocentric visual perception and a Jetson Orin NX (16~GB) as the onboard computing unit, which is responsible for data reception, communication forwarding, and interface management with the low-level flight controller. It should be noted that high-level model inference is performed on the ground-station server, while the UAV mainly handles visual observation acquisition, control command reception, and flight execution. Low-level state estimation and flight control are managed by a CUAV PX4 flight controller operating under closed-loop position control. For reliable data transmission, the UAV connects wirelessly to a local router, which is further tethered to the ground-station server via a wired network, forming a real-time communication link between the UAV and the server.

We designed two real-world experimental scenarios, including indoor and outdoor settings, to comprehensively evaluate the system performance. The indoor experiments were conducted in a 10~m $\times$ 15~m $\times$ 3~m flight arena equipped with a 14-camera motion capture system, which provides highly accurate external pose estimation with sub-millimeter accuracy ($<1$~mm). The pose data from the motion capture system is streamed to the UAV over WiFi and integrated into the PX4 flight controller through the VRPN package and the MAVLink protocol for closed-loop position control and trajectory recording.

The outdoor experiments were conducted in a relatively enclosed, open area with reliable GPS signal reception. To improve the robustness of outdoor state estimation, the GPS data is further complemented by a rigidly mounted Northwake TFmini-S LiDAR rangefinder, which provides accurate altitude estimation. Through the combined evaluation in the indoor motion-capture environment and the outdoor GPS/LiDAR environment, we verify the executability and environmental adaptability of the proposed method on a real UAV platform.

Importantly, neither the motion-capture poses in indoor experiments nor the GPS/LiDAR measurements in outdoor experiments are provided to the model as input; they are used only for low-level flight stabilization, safe motion execution, and trajectory recording, while WorldVLN makes high-level navigation decisions solely from egocentric RGB observations and language instructions.